\documentclass{ifacconf}

\usepackage{graphicx}      
\usepackage{natbib}        

\usepackage{amsmath,bm,amssymb,array }

\usepackage{xcolor}

\newcommand{\tabincell}[2]{\begin{tabular}{@{}#1@{}}#2\end{tabular}}

\DeclareMathOperator{\diag}{diag}
\DeclareMathOperator{\tr}{tr}

\begin{document}
\begin{frontmatter}

\title{Fault Tolerant Control of Mecanum Wheeled Mobile Robots} 


\thanks[footnoteinfo]{Xuehui Ma is with Xi'an University of Technology (xuehui.yx@gmail.com).\\ $^\dag$ Shiliang Zhang is with University of Oslo (shilianz@ifi.uio.no).\\ $^\ddag$ Zhiyong Sun is with Peking University (zhiyong.sun@pku.edu.cn).}

\author{Xuehui Ma$^\star$} \qquad \author{Shiliang Zhang$^\dag$} \qquad \author{Zhiyong Sun$^\ddag$}


\begin{abstract}                
Mecanum wheeled mobile robots (MWMRs) are highly susceptible to actuator faults that degrade performance and risk mission failure. Current fault tolerant control (FTC) schemes for MWMRs target complete actuator failures like motor stall, ignoring partial faults \textit{e.g.}, in torque degradation. We propose an FTC strategy handling both fault types, where we adopt posterior probability to learn real-time fault parameters. We derive the FTC law by aggregating probability-weighed control laws corresponding to predefined faults. This ensures the robustness and safety of MWMR control despite varying levels of fault occurrence. Simulation results demonstrate the effectiveness of our FTC under diverse scenarios.
\end{abstract}


\end{frontmatter}

\section{Introduction}

Mobile robots have been increasingly deployed in industrial, logistics, and medical environments to perform material handling and transportation tasks in recent years (\cite{tzafestas2018mobile}, \cite{zhang2024fault}). Among these platforms, Mecanum wheeled mobile robots (MWMRs) are specifically favored for their holonomic capability, allowing instantaneous and omnidirectional movement (\cite{10607843}, \cite{watson2020collinear}). This agility is critical for superior navigation and smoother motion in confined spaces. However, such agility depends on the precise superposition of forces produced by all individual wheel actuators. This design renders the robot performance sensitive to any control imbalance. Consequently, the occurrence of actuator faults immediately breaks the force balance, severely degrading trajectory tracking and risking mission failures. Actuator faults are a common reality, stemming from factors like motor torque degradation or stall, Mecanum roller wear or jamming, and abrupt changes in ground traction (\textit{e.g.}, slippery surfaces). Therefore, the development of a fault tolerant control (FTC) strategy is essential to maintain functional mobility and ensure safety for MWMRs in the presence of actuator faults. 

Existing FTC schemes are typically classified into two major categories: passive and active FTCs (\cite{zhang2008bibliographical, jiang2012fault,liu2025adaptive,liu2021reliable}). 
Passive FTC utilizes robust control methods (\textit{e.g.}, $H_\infty$ control, sliding mode control) to design controllers resilient to a predefined bounded fault set (\cite{hu2012fault,riaz2021review}). The main drawback of this approach is its over-conservatism. Since robust control is designed to maintain stability for the worst-case fault, it inherently sacrifices control performance for the sake of robustness, even in fault-free scenarios(\cite{ma2024robust,ma2024adaptive}). Active FTC releases this limitation by detecting or learning faults in real-time and dynamically tailoring the control law (\cite{wang2022design,guo2024adaptive,liu2022dual,liu2025adaptive2}). In this way, active FTC reduces conservatism and enhances control accuracy by adapting to actual fault states. 

Recently, several studies have developed active FTC schemes for MWMRs. \cite{karras2020model} and \cite{yang2024extended} designed fault detection to identify the faulty wheels, and then applied model predictive control to derive the FTC control law with the identified faults. \cite{mishra2019behavioural} reconfigured the control allocation using the pseudo-inverse after faults are detected. \cite{alshorman2020fuzzy} combined the fault detection and fuzzy controller to keep the stability and safety of wheeled mobile robots. However, the success of fault detection depends on the accurate measurement of motor torque, which is not available in most low-cost motors, thus limiting their application in real-world robots. To address the limitation in fault detection, \cite{7487389,jin2019adaptive,kamel2017fault} utilized observations like the speed of the robot to learn unknown parameters in MWMRs' dynamics stemming from faults, and employed adaptive control techniques to design the FTC control law. Nevertheless, their adaptive approach addresses scenarios involving only complete actuator failures (\textit{e.g.}, motor stall and jamming, representing a $100\%$ actuator failure). They neglect the more practical scenario of partial actuator failures. In real-world operation, it is common that the long-time working of motors causes torque degradation, resulting in a partial actuator failure (\textit{e.g.}, $50\%$ loss of force). Thus, developing an FTC scheme capable of robustly handling both complete and partial actuator faults is essential. 

In this work, we aim to address this gap by developing an active FTC strategy for \textit{four-Mecanum-wheeled mobile robots} (FMWMRs) that is effective in the presence of complete and partial actuator faults. The main contributions of this work are summarized as follows.
\begin{enumerate}
  \item  We construct a finite set of fault parameters to represent both complete and partial fault parameters. We then explicitly model the actuator faults by incorporating FMWMR driving force and the fault parameters.  
  \item We introduce the posterior probability updating algorithm to learn the true fault parameters from the predefined set in real-time. We also provide the convergence proof for this learning algorithm. 
  \item We design an active FTC law by summing the control law under each predefined fault parameter with the learned posterior probabilities as weights. The probability-weighted summation smooths the control law switching with abrupt faults, thereby ensuring the robustness and safety of the mobile robot. 
\end{enumerate}

The remainder of this paper is as follows. Section~\ref{sec:2} formulates the fault tolerant control problem for FMWMRs. Section~\ref{sec:3} details our FTC strategy and proves the convergence proof for our fault parameter learning. We demonstrate the efficiency of our FTC through numerical simulations in Section~\ref{sec:4}. Finally, Section~\ref{sec:5} concludes this study.

\section{Problem Formulation}\label{sec:2}

In this section, we present the discrete-time dynamic and kinematic model of four-Mecanum-wheeled mobile robots (FMWMRs). Then, we formulate the fault tolerant control problem for FMWMRs with the occurrence of complete and partial actuator faults.  

\subsection{The discrete-time model of FMWMRs}

\begin{figure}
\begin{center}
\includegraphics[width=7.5cm]{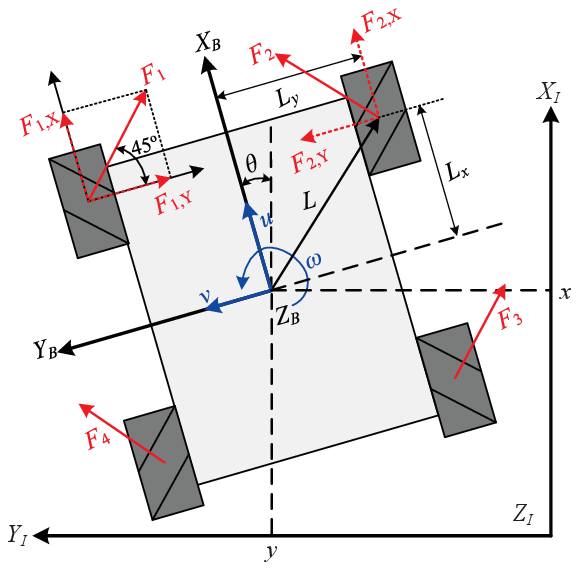}   
\caption{Coordinate frames, wheels configuration and posture definition for FMWMR.} 
\label{fig:MecanumWheel}
\end{center}
\end{figure}

We define the inertial coordinate frame for FMWMR motion as $\{ X_I,Y_I, Z_I\}$ and define the body coordinate frame of FMWMR as $\{X_B, Y_B, Z_B\}$, shown in Fig.~\ref{fig:MecanumWheel}. The continuous-time models for FMWMRs have been established by \cite{alakshendra2017adaptive} and \cite{7487389}. To derive our FTC control, we utilize the forward Euler method to discretize these models with the sampling time interval $T_s$. The resulting discrete-time kinematic model for FMWMR is 
\begin{equation}\label{eq:kinematicmodel}
    \begin{split}
    x_{k+1}&=x_k+T_s(u_k\cos\theta_k-\upsilon_k\sin\theta_k), \\
    y_{k+1}&=y_k+T_s(u_k\sin\theta_k+\upsilon_k\cos\theta_k), \\
    \theta_{k+1}&= \theta_k+T_s\omega_k,
    \end{split} 
\end{equation}
where ($x_k$, $y_k$) and $\theta_k$ represent the position and yaw angle of the robot’s center of mass in the inertial frame, respectively; $u_k$ and $v_k$ are the velocity along the $X_B$-axis and $Y_B$-axis of the body frame, respectively; $\omega_k$ is the rotation rate around the $Z_B$ axis.

The discrete-time dynamic model of FMWMR is 
\begin{equation}\label{eq:dynamicmodel}
    \begin{aligned}
    u_{k+1} &= u_k+T_s\left(\frac{1}{m}(F_{x,k}-c_vu_k)+\upsilon_k\omega_k \right), \\
    \upsilon_{k+1} &= \upsilon_k+T_s\left(\frac{1}{m}(F_{y,k}-c_v\upsilon_k)-u_k\omega_k \right), \\
    \omega_{k+1} &= \omega_k+T_s\left(\frac{1}{I_z}(\tau_{z,k}-c_{\theta}\omega_k) \right),
    \end{aligned} 
\end{equation}
where $m$ is the quality of the robot, $I_z$ is the inertia moment on the $Z_B$ axis. $c_v$ and $c_{\theta}$ are the linear viscous and rotational damping coefficients of the robot. $F_{x,k}$, $F_{y,k}$ and $\tau_{z,k}$ are the resultant force and moment acting on the robot in the body frame, which are calculated by 
\begin{equation}\label{eq:resultantforce}
    \begin{aligned}
        \left[ \begin{array}{c} F_{x,k} \\ F_{y,k} \\ \tau_{z,k} \end{array} \right] =  \frac{1}{\sqrt{2}} \left[ \begin{array}{cccc} 1 & 1 & 1 & 1 \\ -1 & 1 & -1 & 1 \\ -\bar{L} & \bar{L} & \bar{L} & -\bar{L} \end{array} \right] \left[ \begin{array}{c} F_{1,k} \\ F_{2,k} \\ F_{3,k} \\ F_{4,k} \end{array} \right], 
    \end{aligned} 
\end{equation}
where the parameter $\bar{L}$ is $ (L_x+L_y) $. $F_{i,k}$ is the driving force generated by the $i$-th wheel. The driving force is derived from
\begin{equation}\label{eq:drivingforce}
F_{i,k} = (\tau_{i,k}-\tau_{f,i})/r,
\end{equation}
where $\tau_{i,k}$ is the driving torque of wheel $i$, and $\tau_f$ is the friction torque acting on the wheel, including rolling resistance, Coulomb friction, and viscous damping effects. 

\subsection{The fault tolerant control problem formulation}\label{subsectionprobform}

In this work, we aim to develop a robust control strategy for FMWMRs to track a reference trajectory, despite the presence of actuator faults. We model the actuator faults by multiplying the driving force by a designated fault parameter, and then the driving force becomes
\begin{equation}\label{eq:faultforce}
F_{i,k} =  (\lambda_{i}\tau_{i,k}-\tau_{f,i})/r,
\end{equation}
where $\lambda_{i}$ is the fault parameter, representing the operational status for actuator $i$. 

The redundancy of wheel-actuators allows FMWMRs to retain actuation capability with one failed wheel actuator. This means the FMWMR can still be fully controlled with three healthy wheel actuators. 
However, the FMWMR becomes underactuated when two actuators fail simultaneously, as illustrated in \cite{7487389}.  We in this study aim to develop an FTC to ensure that the FMWMR maintains safety and collision avoidance in both one-fault and two-fault cases. To cover both faults, we define three fault scenarios for the parameter vector $\Lambda = [\lambda_1,\lambda_2,\lambda_3,\lambda_4]$, as explained below. Note that the control of FMWMRs will be completely impeded when three or more actuators fail.

\begin{enumerate}
  \item Fault-free case. All four actuators are functioning at full capacity. We set the fault parameters as $\Lambda^1 = [1,1,1,1]$, where the value $\lambda_i = 1$ signifies $100\%$ operational status for actuator $i$; 
  \item One-fault case. One actuator is in complete or partial failure. The value $\lambda_i = 0$ signifies a $100\%$ loss of capacity for actuator $i$. We set the complete fault parameters for one failed wheel as $\Lambda^2 = [0,1,1,1]$, $\Lambda^3 = [1,0,1,1]$, $\Lambda^4 = [1,1,0,1]$, $\Lambda^5 = [1,1,1,0]$. In addition, the value $\lambda_i = 0.5$ signifies a $50\%$ loss of capacity for actuator $i$. Similarly, we set the fault parameters with 50\% partial failure as $\Lambda^6 = [0.5,1,1,1]$, $\Lambda^7 = [1,0.5,1,1]$, $\Lambda^8 = [1,1,0.5,1]$, $\Lambda^9 = [1,1,1,0.5]$.  
  \item Two-fault case. We set the parameter vector with $100\%$ loss of actuator capacity as $\Lambda^{10} = [0,0,1,1]$, $\Lambda^{11} = [1,0,0,1]$, $\Lambda^{12} = [1,1,0,0]$, $\Lambda^{13} = [0,1,1,0]$. We set the parameter vector with $50\%$ loss of actuator capacity as $\Lambda^{14} = [0.5,0.5,1,1]$, $\Lambda^{15} = [1,0.5,0.5,1]$, $\Lambda^{16} = [1,1,0.5,0.5]$, $\Lambda^{17} = [0.5,1,1,0.5]$.
\end{enumerate}

Note that existing FTC schemes often simplify actuator failures and only consider the $100\%$ actuator fault case, while ignoring partial loss of actuator capacity. Here, we introduce finer partial failure parameters (\textit{e.g.}, $25\%$, $50\%$, $75\%$, \textit{etc}.) that would offer a more precise representation of faults. 
Note that considering multiple levels of partial failure increases the fault set size, which leads to additional computation in learning the fault type and calculating the control law. Therefore, we only take the value $50\%$ for partial failure cases to balance the requirement of control accuracy and computational efficiency. 

With the predefined fault parameter set, we formulate the fault tolerant tracking control problem as follows:
\begin{equation}\label{MPC_P}
    \begin{aligned}
    (P) \quad &\min_{\bm{u}_k} \quad J_k \\
    \text{s.t.} \quad & (1), (2), (3), (5), \\
                      & \Lambda \in \{ \Lambda^1, \Lambda^2, \cdots, \Lambda^{s} \}.
    \end{aligned} 
\end{equation}
where the fault parameter set $\Lambda^i$ is user-defined, and $s$ is the element number. In solving problem $(P)$, we adopt the model predictive control (MPC) framework and we define the cost function as
\begin{equation} \label{costfcn_P}
\begin{aligned}
J_k & = \| \bm{x}_{k+N}-\bm{x}_{d,k+N} \|^2_{\bm{Q}_{k+N}} \\
&+\sum_{i=0}^{N-1} \left( \| \bm{x}_{k+i}-\bm{x}_{d,k+i} \|^2_{\bm{Q}_{k+i}} + \| \bm{u}_{k+i} \|^2_{\bm{R}_{k+i}} \right), \\
\end{aligned} 
\end{equation}
where the state vector is defined as $\bm{x}_k = [x_k, y_k, \omega_k]^T$, the control vector is defined as $\bm{u}_k = [\tau_{1,k}, \tau_{2,k}, \tau_{3,k}, \tau_{4,k}]^T$, $\bm{x}_{d,k}$ is the desired trajectory, $N$ is the prediction horizon in MPC. In the next section, we explain how to solve this control problem and derive the fault tolerant control law.

\section{Fault Tolerant Control Design}\label{sec:3}

\begin{figure}
\begin{center}
\includegraphics[width=9cm]{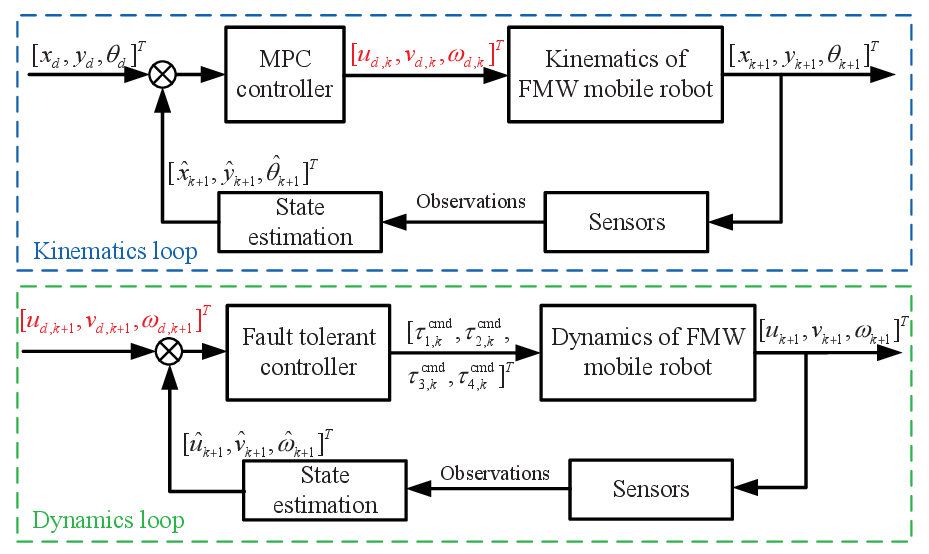}    
\caption{The diagram of the kinematics control loop and dynamics control loop in the fault tolerant control of FMWMRs.} 
\label{fig:controlloop}
\end{center}
\end{figure}

This section describes the derivation of our fault tolerant control law. As is shown in Fig.~\ref{fig:controlloop}, we divide the control structure into two loops: (i) Kinematics loop, and (ii) Dynamics loop. Following this structure, we separate the MPC problem $(P)$ presented in Section~\ref{sec:2} into new control problems: (i) MPC problem in kinematics loop $(P_{\text{kine}})$, and (ii) fault tolerant control problem $(P_{\text{dyna}})$ in dynamics loop. Below we elaborate the solutions for the two problems in detail.

\subsection{MPC in the kinematics loop}\label{sectrionKineloop}
In this section, we firstly employ extended Kalman filter (EKF) to estimate the state of FMWMR. We define the state vector as $\bm{x}_k = [x_k, y_k, \omega_k]^T$, representing the position and orientation of robots. To simplify the notation of the kinematic model in (\ref{eq:kinematicmodel}), we introduce the function $F_{\text{kine}}(\cdot)$. We utilize the EKF to fuse the measurements from different sensors for estimating $\bm{x}_k$. We account for process noise $\bm{e}_{\text{process}}$ in state evolution, and observation noise $\bm{e}_{\text{observe}}$ in observation update, as shown below:
\begin{equation}\label{eq:stochastickinematics}
\begin{aligned}
&\bm{x}_{k+1} = F_{\text{kine}}(\bm{x}_k,\bm{\xi}_{k})+\bm{e}_{\text{process},k} \\
&\bm{x}_{\text{obs},k} = \bm{x}_k + \bm{e}_{\text{observe},k} \\
\end{aligned} 
\end{equation}
where $\bm{x}_{\text{obs},k}$ is the state observation, $\bm{\xi}_{k}=[u_{k},v_{k},\omega_{k}]^T$ is the control vector. The process and observation noises are assumed to follow Gaussian distributions, represented as $\bm{e}_{\text{process},k} \sim \mathcal{N}(0, \bm{Q}_{\text{kine},k})$ and $\bm{e}_{\text{observe},k} \sim \mathcal{N}(0, \bm{R}_{\text{kine},k})$. 

The EKF-based state estimation for position and orientation of the robot motion is 
\begin{equation}
\begin{aligned}
    &\hat{\bm{x}}_{k|k-1} = F_{\text{kine}}(\hat{\bm{x}}_{k-1|k-1}, \bm{\xi}_{k-1}), \\
    &\bm{P}_{k|k-1} = \bm{F}_{\text{kine},k}\bm{P}_{k-1|k-1}\bm{F}^T_{\text{kine},k} + \bm{Q}_{\text{kine},k-1},\\
    &\bm{K}_k = \bm{P}_{k|k-1} \bm{H}^T_{\text{kine},k} (\bm{H}_{\text{kine},k}\bm{P}_{k|k-1} \bm{H}^T_{\text{kine},k} + \bm{R}_{\text{kine},k} )^{-1},  \\
    &\hat{\bm{x}}_{k|k} = \hat{\bm{x}}_{k|k-1} + \bm{K}_k( \bm{x}_{\text{obs},k} - \bm{H}_{\text{kine},k}\hat{\bm{x}}_{k|k-1}), \\
    &\bm{P}_{k|k} = (\bm{I} - \bm{K}_k\bm{H}_{\text{kine},k}) \bm{P}_{k|k-1},
\end{aligned} 
\end{equation}
where the state transition matrix and observation matrix are
\begin{equation}
\begin{aligned}
\bm{F}_{\text{kine},k} = \left . \frac{\partial F_{\text{kine}}(\cdot)}{\partial \bm{x} } \right|_{\hat{\bm{x}}_{k-1|k-1},\bm{\xi}_{k-1}}, \bm{H}^T_{\text{kine},k} = \bm{I}_{3\times3}.
\end{aligned} 
\end{equation}

We obtain the control signal $\bm{\xi}_{k}^{\ast}$ that drives the robot tracking the desired trajectory $\bm{x}_{d,k}$ by solving the following MPC control problem:
\begin{equation}
\begin{aligned}
(P_{\text{kine}}) \quad &\min_{\bm{\xi}_{k}} \quad J_{\text{kine},k} \\
    \text{s.t.} \quad  &(8),  \quad \bm{\xi}_{k} \in [\bm{\xi}_{\text{min}},\bm{\xi}_{\text{max}}],
\end{aligned} 
\end{equation}
where the cost function is defined as
\begin{equation}
\begin{aligned}
& J_{\text{kine},k} = \| \bm{x}_{k+N}-\bm{x}_{d,k+N} \|^2_{\bm{Q}_{k+N}} \\
&+\sum_{i=0}^{N-1} \left( \| \bm{x}_{k+i}-\bm{x}_{d,k+i} \|^2_{\bm{Q}_{k+i}} + \| \bm{\xi}_{k+i} \|^2_{\bm{R}_{k+i}} \right). \\
\end{aligned} 
\end{equation}
In this work, we linearize the kinematic model (\ref{eq:stochastickinematics}), and transfer the nonlinear MPC problem $(P_{\text{kine}})$ into a standard linear MPC problem. We then employ a QP solver (\textit{e.g.}, OSQP) (\cite{osqp}) to solve this linear MPC control problem, to obtain the control signal $\bm{\xi}_{k}$. The solver OSQP is computationally friendly and efficient for resource-constrained embedded systems.

\subsection{Fault tolerant control in the dynamics loop}
In this section, we design an adaptive fault tolerant control strategy in the dynamics loop for robots with one or two wheels in failure. To simplify the notation of the dynamic model, we combine the dynamic model (\ref{eq:dynamicmodel}), resultant force and moment (\ref{eq:resultantforce}), and wheel driving force (\ref{eq:drivingforce}) to one equation, shown as
\begin{equation}\label{eq:stochasticdynamics}
\bm{\xi}_{k+1} = F_{\text{dyna}}(\bm{\xi}_{k}) + \bm{G}_{\text{dyna}} \bm{u}_k + \bm{\varepsilon}_{\text{process},k}
\end{equation}
where the control matrix is
\begin{equation}
\begin{aligned}
\bm{G}_{\text{dyna}} = &\frac{T_s}{\sqrt{2}r} \left[ \begin{array}{ccc} \frac{1}{m} & 0 & 0 \\ 0 & \frac{1}{m} & 0 \\ 0 & 0 & \frac{1}{I_z} \end{array} \right]  \left[ \begin{array}{cccc} 1 & 1 & 1 & 1 \\ -1 & 1 & -1 & 1 \\ -\bar{L} & \bar{L} & \bar{L} & -\bar{L} \end{array} \right] \diag(\Lambda),
\end{aligned} 
\end{equation}
and the state matrix is 
\begin{equation}
\begin{aligned}
F_{\text{dyna}}(\bm{\xi}_{k}) =  \left[ \begin{array}{c} u_k \\ v_k \\ \omega_k \end{array} \right] + T_s \left[ \begin{array}{c}  -\frac{c_vu_k}{m}+v_k\omega_k \\  -\frac{c_vv_k}{m}-u_k\omega_k \\  -\frac{c_{\theta}\omega_k}{I_z} \end{array}\right] \\ 
+\frac{T_s}{\sqrt{2} r}\left[ \begin{array}{ccc} \frac{1}{m} & 0 & 0 \\ 0 &\frac{1}{m} & 0 \\ 0 & 0 & \frac{1}{I_z} \end{array} \right]  \left[ \begin{array}{cccc} 1 & 1 & 1 & 1 \\ -1 & 1 & -1 & 1 \\ -\bar{L} & \bar{L} & \bar{L} & -\bar{L} \end{array} \right] \left[ \begin{array}{c} \tau_{f,1}\\ \tau_{f,2} \\ \tau_{f,3}\\ \tau_{f,4} \end{array} \right].
\end{aligned} 
\end{equation}
We also consider the process noise, which is assumed to follow the distribution $\bm{\varepsilon}_{\text{process},k} \sim \mathcal{N}(0, \bm{Q}_{\text{dyna},k}) $. The observation equation is 
\begin{equation}\label{eq:stochasticdynamicsobs}
\bm{\xi}_{\text{obs},k} = \bm{\xi}_{k} + \bm{\varepsilon}_{\text{observe},k}
\end{equation}
where the observation noise is assumed to be Gaussian $\bm{\varepsilon}_{\text{observe},k} \sim \mathcal{N}(0, \bm{R}_{\text{dyna},k})$. 

As mentioned in Section~\ref{subsectionprobform}, the parameter $\Lambda$ in the state matrix $\bm{G}_{\text{dyna}}$ gets values from the finite set $\{ \Lambda^1, \Lambda^2, \cdots, \Lambda^{s} \}$. Hence the matrix $\bm{G}_{\text{dyna}}$ belongs to the finite set $\{ \bm{G}_{\text{dyna}}^1, \bm{G}_{\text{dyna}}^2, \cdots, \bm{G}_{\text{dyna}}^{s} \}$.
Then, we present the fault tolerant control problem of velocity and angular rate of the robot motion as:
\begin{equation}
\begin{aligned}
(P_{\text{dyna}}) \quad &\min_{\bm{u}_{k}} \quad J_{\text{dyna},k} \\
    \text{s.t.} \quad  &(14),  \quad \bm{u}_{k} \in [\bm{u}_{\text{min}},\bm{u}_{\text{max}}], \\
    & \bm{G}_{\text{dyna}} \in \{ \bm{G}_{\text{dyna}}^1, \bm{G}_{\text{dyna}}^2, \cdots, \bm{G}_{\text{dyna}}^{s}\},
\end{aligned} 
\end{equation}
where the cost function is defined as
\begin{equation}
\begin{aligned}
J_{\text{dyna},k} = E \left\{ \left. \| \bm{\xi}_{k+1}-\bm{\xi}_{d,k+1} \|^2 + \beta\| \bm{u}_{k} \|^2 \right| \Xi_k \right\} , \\
\end{aligned} 
\end{equation}
where $\Xi_k$ is the historical measured information 
\begin{equation}
\Xi_k = [\bm{\xi}_{\text{obs},0},\bm{\xi}_{\text{obs},1}, \cdots, \bm{\xi}_{\text{obs},k}]^T.
\end{equation}
The desired velocity and angular rate represent the optimal solution obtained from Section \ref{sectrionKineloop}, denoted as $\bm{\xi}_{d,k+1} = \bm{\xi}_{k+1}^{\ast}$. 
The calculation for minimizing the cost in problem $(P_{\text{dyna}})$ can be rewritten as:
\begin{equation}
\begin{aligned}
&\min_{\bm{u}_{k}}  E \left\{ \left. \| \bm{\xi}_{k+1}-\bm{\xi}_{d,k+1} \|^2 + \beta\| \bm{u}_{k} \|^2 \right| \Xi_k \right\} \\
\Rightarrow & \min_{\bm{u}_{k}}  E \left\{  E \left\{ \| \bm{\xi}_{k+1}-\bm{\xi}_{d,k+1} \|^2 \right. \right.\\
& \quad \quad \quad \quad \quad\left. \left. \left. \left. + \beta\| \bm{u}_{k} \|^2 \right| \Xi_k,\bm{G}_{\text{dyna}}^i \right\} \right| \Xi_k \right\} \\
\Rightarrow &  E \left\{ \min_{\bm{u}_{k}} E \left\{ \| \bm{\xi}_{k+1}-\bm{\xi}_{d,k+1} \|^2 \right. \right.\\
& \quad \quad \quad \quad \quad\left. \left. \left. \left. + \beta\| \bm{u}_{k} \|^2 \right| \Xi_k,\bm{G}_{\text{dyna}}^i \right\} \right| \Xi_k \right\} \\
\Rightarrow & \sum_{i=1}^{s} \pi(\bm{G}_{\text{dyna}}^i|\bm{\Xi}(k)) \min_{\bm{u}_{k}} E \left\{ \| \bm{\xi}_{k+1}-\bm{\xi}_{d,k+1} \|^2 \right. \\
& \quad \quad \quad \quad \quad \left. \left. + \beta\| \bm{u}_{k} \|^2 \right| \Xi_k,\bm{G}_{\text{dyna}}^i \right\} 
\end{aligned} 
\end{equation}
Hence the optimal control law $\bm{u}_{k}^{\ast}$ is 
\begin{equation}\label{optctrllaw}
\bm{u}_{k}^{\ast} = \sum_{i=1}^{s}\pi(\bm{G}_{\text{dyna}}^i|\bm{\Xi}_k) \bm{u}_{k|\bm{G}_{\text{dyna}}^i},
\end{equation}
where the control signal $\bm{u}_{k|\bm{G}_{\text{dyna}}^i}$ is derived by solving the following optimization problem 
\begin{equation}
\begin{aligned}
(P_{\text{dyna}}^i) \quad &\min_{\bm{u}_{k}} \quad J_{\text{dyna},k}^i \\
    \text{s.t.} \quad  &\bm{\xi}_{k+1} = F_{\text{dyna}}(\bm{\xi}_{k}) + \bm{G}_{\text{dyna}}^i \bm{u}_k + \bm{\varepsilon}_{\text{process},k}\\
    &\bm{\xi}_{\text{obs},k} = \bm{\xi}_{k} + \bm{\varepsilon}_{\text{observe},k} \\
    &\bm{u}_{k} \in [\bm{u}_{\text{min}},\bm{u}_{\text{max}}], \\
\end{aligned} 
\end{equation}
where the cost function is
\begin{equation}
J_{\text{dyna},k}^i = E \left\{ \left. \| \bm{\xi}_{k+1}-\bm{\xi}_{d,k+1} \|^2 + \beta\| \bm{u}_{k} \|^2 \right| \Xi_k,\bm{G}_{\text{dyna}}^i \right\}.
\end{equation}
Following the optimality condition  $\frac{\partial J_{\text{dyna},k}^i }{\partial \bm{u}_{k}} = 0$, we obtain the control law
\begin{equation}\label{controllaw_i}
\begin{aligned}
\bm{u}_{k|\bm{G}_{\text{dyna}}^i} =& -(\bm{G}_{\text{dyna}}^{i,T}\bm{G}_{\text{dyna}}^{i} + \beta\bm{I}_{4\times4})^{-1} \\&(F_{\text{dyna}}^T(\hat{\bm{\xi}}_{k|k,\bm{G}_{\text{dyna}}^{i}}) - \bm{\xi}_{d,k+1} ) \bm{G}_{\text{dyna}}^{i},
\end{aligned} 
\end{equation}
where $\hat{\bm{\xi}}_{k|k,\bm{G}_{\text{dyna}}^{i}}$ is the estimation of state vector $\bm{\xi}_{k}$, which is derived based on the EKF. The estimation algorithm of velocity and angular rate of the FMWMR with respect to each of the parameter matrix $\bm{G}_{\text{dyna}}^i$ is presented as 
\begin{equation}
\begin{aligned}
    &\hat{\bm{\xi}}_{k|k-1,\bm{G}_{\text{dyna}}^i} = F_{\text{dyna}}(\hat{\bm{\xi}}_{k-1|k-1,\bm{G}_{\text{dyna}}^i})+ \bm{G}_{\text{dyna}}^i\bm{u}_{k-1}, \\
    &\bm{\Sigma}_{k|k-1,\bm{G}_{\text{dyna}}^i} = \bm{F}_{\text{dyna},k}\bm{\Sigma}_{k-1|k-1,\bm{G}_{\text{dyna}}^i}\bm{F}^T_{\text{dyna},k} + \bm{Q}_{\text{dyna},k-1},\\
    &\bm{K}_{k|\bm{G}_{\text{dyna}}^i} = \bm{\Sigma}_{k|k-1,\bm{G}_{\text{dyna}}^i} \bm{H}^T_{\text{dyna},k} \\
    &\quad\quad\quad\quad\quad(\bm{H}_{\text{dyna},k}\bm{\Sigma}_{k|k-1,\bm{G}_{\text{dyna}}^i} \bm{H}^T_{\text{dyna},k} + \bm{R}_{\text{dyna},k} )^{-1},  \\
    &\hat{\bm{\xi}}_{k|k,\bm{G}_{\text{dyna}}^i} = \hat{\bm{\xi}}_{k|k-1,\bm{G}_{\text{dyna}}^i} + \bm{K}_{k|\bm{G}_{\text{dyna}}^i} \\
    &\quad\quad\quad\quad\quad( \bm{\xi}_{\text{obs},k} - \bm{H}_{\text{dyna},k}\hat{\bm{\xi}}_{k|k-1,\bm{G}_{\text{dyna}}^i}), \\
    &\bm{\Sigma}_{k|k,\bm{G}_{\text{dyna}}^i} = (\bm{I} - \bm{K}_{k|\bm{G}_{\text{dyna}}^i}\bm{H}_{\text{dyna},k}) \bm{\Sigma}_{k|k-1,\bm{G}_{\text{dyna}}^i},
\end{aligned} 
\end{equation}
where the state transition matrix and observation matrix are
\begin{equation}
\begin{aligned}
\bm{F}_{\text{dyna},k} = \left . \frac{\partial F_{\text{dyna}}(\cdot)}{\partial \bm{\xi} } \right|_{\hat{\bm{\xi}}_{k-1|k-1},\bm{u}_{k-1}}, \bm{H}^T_{\text{dyna},k} = \bm{I}_{3\times3}.
\end{aligned} 
\end{equation}

The Bayesian posterior probability in the control (\ref{optctrllaw}) is updated by 
\begin{equation}\label{eq:postprobupdate}
\begin{aligned}
\pi(\bm{G}_{\text{dyna}}^i|\bm{\Xi}_k) = \frac{p(\bm{\xi}_k|\bm{G}_{\text{dyna}}^i,\bm{\Xi}_{k-1}) \pi(\bm{G}_{\text{dyna}}^i|\bm{\Xi}_{k-1})}{\sum_{j=1}^{s}p(\bm{\xi}_k|\bm{G}_{\text{dyna}}^j,\bm{\Xi}_{k-1})\pi(\bm{G}_{\text{dyna}}^j|\bm{\Xi}_{k-1})},
\end{aligned} 
\end{equation}
where the likelihood $p(\bm{\xi}_k|\bm{G}_{\text{dyna}}^i,\bm{\Xi}_{k-1})$ is calculated by
\begin{equation}
\begin{aligned}
&p(\bm{\xi}_k|\bm{G}_{\text{dyna}}^i,\bm{\Xi}_{k-1})\\
= &\left| \Sigma_{\text{obs},k}^i \right|^{-\frac{1}{2}} \exp\left\{ -\frac{1}{2}(\tilde{\bm{\xi}}_{k|k-1}^i)^T(\Sigma_{\text{obs},k}^i)^{-1}\tilde{\bm{\xi}}_{k|k-1}^i \right\}, \\
\end{aligned} 
\end{equation}
where the observation covariance matrix is
\begin{equation}
\Sigma_{\text{obs},k}^i = \bm{H}_{\text{dyna},k}\bm{\Sigma}_{k|k-1,\bm{G}_{\text{dyna}}^i} \bm{H}^T_{\text{dyna},k} + \bm{R}_{\text{dyna},k}, 
\end{equation}
and the innovation is 
\begin{equation}
\tilde{\bm{\xi}}_{k|k-1}^i = \bm{\xi}_{\text{obs},k} - \bm{H}_{\text{dyna},k}\hat{\bm{\xi}}_{k|k-1,\bm{G}_{\text{dyna}}^i}.
\end{equation}

Note that combining the analytical solution in (\ref{controllaw_i}) and the posterior probability in (\ref{eq:postprobupdate}), the control law in (\ref{optctrllaw}) has an explicit form, thus leading to a computationally friendly solution for resource-constrained embedded systems.

\subsection{The convergence proof of fault parameter learning}
This section proves the convergence of our parameter learning algorithm. The algorithm depends on the posterior probability update (\ref{eq:postprobupdate}) for each fault parameter vector within the predefined set $\{ \bm{G}_{\text{dyna}}^1, \bm{G}_{\text{dyna}}^2, \cdots, \bm{G}_{\text{dyna}}^{s} \}$. Our convergence proof demonstrates that the posterior probability corresponding to true parameter, $\pi(\bm{G}_{\text{dyna}}^i|\bm{\Xi}_k)$, converges to $1$, while the probabilities for other (incorrect) parameters, $\pi(\bm{G}_{\text{dyna}}^j|\bm{\Xi}_k)$ for $ j\neq i$, converge to $0$. We first make a statement in Lemma \ref{lemma1} that underpins our convergence later, and then we provide the full convergence proof for the posterior probability in Theorem \ref{thm1}.

\begin{lem} \label{lemma1}
If $\bm{A}\in \mathbb{R}^{l\times l} $ and $\bm{B}\in \mathbb{R}^{l\times l}$ are positive definite, then 
\begin{equation}\label{eq:lemma1}
l+\ln \left\{ \frac{|\bm{A}|}{|\bm{B}|}\right\}- \tr \left\{ \bm{B}^{-1} \bm{A} \right\} \leq 0,
\end{equation}
with equality if and only if $\bm{A}=\bm{B}$.
\end{lem}
\begin{pf}
We consider the matrix $\bm{C}= \bm{B}^{-\frac{1}{2}}\bm{A}\bm{B}^{-\frac{1}{2}}$. Then, $|\bm{C}|=\frac{|\bm{A}|}{|\bm{B}|}$ and $\tr(\bm{C})=\tr(\bm{B}^{-1}\bm{A})$. Because $\bm{A}\succ0$ and $\bm{B}\succ0$, the matrix $\bm{C}\succ0$. Thus the inequality (\ref{eq:lemma1}) becomes $l+\ln |\bm{C}| -\tr (\bm{C}) \leq 0$. Let the eigenvalues of $\bm{C}$ be $\lambda_i,i=1,2,\cdots,l$, where $\lambda_i>0$. We have $|\bm{C}|=\prod_{i=1}^l\lambda_i$ and $\tr(\bm{C})=\sum_{i=1}^l\lambda_i$. 
Then the inequality can be rewritten as $\sum_{i=1}^l (1+ \ln \lambda_i-\lambda_i) \leq 0$. For $\forall a > 0$, $1+\ln a -a \leq 0 $ holds, with equality iff $a=1$. Thus, we have $1+ \ln \lambda_i -\lambda_i \leq 0$ for each $i$. By summing over all $i$, we obtain $\sum_{i=1}^l (1+ \ln \lambda_i-\lambda_i) \leq 0$, which is $l+\ln |\bm{C}| -\tr (\bm{C}) \leq 0$. The equality requires $\bm{C}=\bm{I}$, that is $\bm{A}=\bm{B}$.
\end{pf}

\begin{thm} \label{thm1}
Suppose $\bm{G}_{\text{dyna}}^i$ is the true parameter, the convergence of the probabilities are $\lim_{k\rightarrow\infty}\pi(\bm{G}_{\text{dyna}}^i|\bm{\Xi}_k)=1$ and $\lim_{k\rightarrow\infty}\pi(\bm{G}_{\text{dyna}}^j|\bm{\Xi}_k)=0, j\neq i$.
\end{thm} 
\begin{pf}
Suppose $\bm{G}_{\text{dyna}}^i$ is the true parameter. Let 
\begin{equation}\label{eq:Gammafcn}
\begin{aligned}
\Gamma^j_k = \frac{\pi(\bm{G}_{\text{dyna}}^j|\bm{\Xi}_k)}{\pi(\bm{G}_{\text{dyna}}^i|\bm{\Xi}_k)}, \quad j \neq i. 
\end{aligned} 
\end{equation}
Substituting (\ref{eq:postprobupdate}) into (\ref{eq:Gammafcn}), we get
\begin{equation}\label{eq:Gammafcn2}
\begin{aligned}
&\Gamma^j_k = \frac{\left| \Sigma_{\text{obs},k}^j \right|^{-\frac{1}{2}} \exp\left\{ -\frac{1}{2}(\tilde{\bm{\xi}}_{k|k-1}^j)^T(\Sigma_{\text{obs},k}^j)^{-1}\tilde{\bm{\xi}}_{k|k-1}^j \right\}}{\left| \Sigma_{\text{obs},k}^i \right|^{-\frac{1}{2}} \exp\left\{ -\frac{1}{2}(\tilde{\bm{\xi}}_{k|k-1}^i)^T(\Sigma_{\text{obs},k}^i)^{-1}\tilde{\bm{\xi}}_{k|k-1}^i \right\}}
\Gamma^j_{k-1} \\
\end{aligned} 
\end{equation}
We take the natural logarithms of (\ref{eq:Gammafcn2}) and obtain
\begin{equation}\label{eq:Gammafcn3}
\begin{aligned}
&\ln \left\{ \frac{\Gamma^j_{k+n-1}}{\Gamma^j_{k-1}} \right\} = \frac{1}{2}n\ln \left\{ \frac{|\Sigma_{\text{obs},k}^i|}{|\Sigma_{\text{obs},k}^j|} \right\} \\
&-\frac{1}{2}\tr\left\{ \sum_{t=k}^{k+n-1} \tilde{\bm{\xi}}_{k|k-1}^j(\tilde{\bm{\xi}}_{k|k-1}^j)^T(\Sigma_{\text{obs},k}^j)^{-1} \right\} \\
&+\frac{1}{2}\tr\left\{ \sum_{t=k}^{k+n-1} \tilde{\bm{\xi}}_{k|k-1}^i(\tilde{\bm{\xi}}_{k|k-1}^i)^T(\Sigma_{\text{obs},k}^i)^{-1} \right\} 
\end{aligned} 
\end{equation}
We suppose $\tilde{\bm{\xi}}_{k|k-1}^i$ is a weakly asymptotically stationary sequence, and is ergodic for all $i$. Thus, its covariance $\Sigma_{\text{obs},k}^i$ is constant, and is represented by $\Sigma^i$. We have the limits
\begin{equation}\label{eq:limit1}
\lim_{n\rightarrow\infty}\frac{1}{n} \sum_{t=k}^{k+n-1} \tilde{\bm{\xi}}_{k|k-1}^i(\tilde{\bm{\xi}}_{k|k-1}^i)^T(\Sigma^i)^{-1} =\bm{I},
\end{equation}
and 
\begin{equation}\label{eq:limit2}
\lim_{n\rightarrow\infty}\frac{1}{n} \sum_{t=k}^{k+n-1} \tilde{\bm{\xi}}_{k|k-1}^j(\tilde{\bm{\xi}}_{k|k-1}^j)^T \geq \Sigma^i, \forall j.
\end{equation}
Substituting (\ref{eq:limit1}) and (\ref{eq:limit2}) into (\ref{eq:Gammafcn3}), we obtain
\begin{equation}\label{eq:limit3}
\begin{aligned}
\lim_{n\rightarrow\infty} \frac{2}{n} \ln \left\{ \frac{\Gamma^j_{k+n-1}}{\Gamma^j_{k-1}}\right\} = \ln \left\{ \frac{|\Sigma^i|}{|\Sigma^j|} \right\} 
-\tr \{ (\Sigma^j)^{-1} \Sigma^i\}\\ +\tr\{\bm{I}\} 
-\tr\{(\Sigma^j)^{-1}M_j\}
\end{aligned} 
\end{equation}
where
\begin{equation}\label{eq:Mmatrix}
M_j = \lim_{n\rightarrow\infty} \frac{1}{n} \sum_{t=k}^{k+n-1} \tilde{\bm{\xi}}_{k|k-1}^j (\tilde{\bm{\xi}}_{k|k-1}^j)^T -\Sigma^i
\end{equation}
According to Lemma \ref{lemma1} and (\ref{eq:limit3}), we have 
\begin{equation}
\ln \left\{ \frac{|\Sigma^i|}{|\Sigma^j|} \right\} 
-\tr \{ (\Sigma^j)^{-1} \Sigma^i\} +\tr\{\bm{I}\} \leq 0
\end{equation}
substituting (\ref{eq:limit2}) into (\ref{eq:Mmatrix}), we obtain that the matrix $M_j \succeq 0$. Thus, $\tr\{(\Sigma^j)^{-1}M_j\}\succeq0$, and $\tr\{(\Sigma^j)^{-1}M_j\}\succ0$ iff $\Sigma^j\neq\Sigma^i$. Then we have
\begin{equation}\label{eq:limit4}
\lim_{n\rightarrow\infty} \frac{2}{n} \ln \left\{ \frac{\Gamma^j_{k+n-1}}{\Gamma^j_{k-1}}\right\} = -c , \forall j\neq i,
\end{equation}
where $c>0$. We can rewrite (\ref{eq:limit4}) as 
\begin{equation}\label{eq:limit5}
\lim_{n\rightarrow\infty} \Gamma^j_{k+n-1} = K \exp\{-\frac{nc}{2}\} \Gamma^j_{k-1} , \forall j\neq i,
\end{equation}
where $K$ is a constant. Combining (\ref{eq:Gammafcn}) and (\ref{eq:limit5}) we obtain 
\begin{equation}\label{eq:limit6}
\lim_{k\rightarrow\infty}\frac{\pi(\bm{G}_{\text{dyna}}^j|\bm{\Xi}_k)}{\pi(\bm{G}_{\text{dyna}}^i|\bm{\Xi}_k)}=0,\forall j\neq i,
\end{equation}
With (\ref{eq:limit6}), we get 
\begin{equation}\label{eq:limit7}
\lim_{k\rightarrow \infty} \pi(\bm{G}_{\text{dyna}}^j|\bm{\Xi}_k) =0, \forall j\neq i,
\end{equation}
and 
\begin{equation}\label{eq:limit8}
\lim_{k\rightarrow \infty} \pi(\bm{G}_{\text{dyna}}^j|\bm{\Xi}_k) =1.
\end{equation}

\end{pf}

\section{Simulations and Results}\label{sec:4}
In the simulations, we compare our method with PID controller and the adaptive MPC controller (\cite{7487389}) to demonstrate the efficiency of our method. The parameters of the mobile robot in the simulations are shown in Table~\ref{tb:parameters}. Below we detail the simulations under different fault scenarios.
\begin{table}[hb]\label{table1}
\begin{center}
\caption{FMWMR parameters}\label{tb:parameters}
\begin{tabular}{ccc}
\hline
 Notation & Physical meaning & Value \\ 
\hline
 $m$ & Quality of the robot & $3$ kg \\  
 $I_z$ &  \tabincell{c}{Moment of inertia \\around the $Z_B$-axis} & $1.2$ kg$\cdot \text{m}^2$ \\
 $r$ & Radius of wheel & $0.04$ m \\
 $L_x$ & Half of the length of robot  & $0.1$ m \\
 $L_y$ & Half of the width of robot & $0.1$ m \\
 $c_v$ & Linear damping & $2$ N$\cdot$s/m \\
 $c_{\theta}$ & Angular damping & $0.1$ N$\cdot$m$\cdot$s \\
 $\tau_{f,i}$ & Friction torque & $0.05$ N$\cdotp$m \\
 $\tau_i$ & Driving torque & $[-0.5,0.5]$ N$\cdotp$m \\
 \hline
\end{tabular}
\end{center}
\end{table}

\subsection{One-fault case}
In this case, we test the trajectory tracking performance for FMWMR with one-fault occurrence in the actuators. Here, we incorporate a sequential change in actuator faults, and the fault parameter is set as
\begin{equation}\label{eq:onefault}
	\begin{aligned}
        \Lambda^T (t) = \left\{ \begin{array}{lr} (0.5,0,1,1), &  0\text{s} \leq t < 10\text{s} ,\\ (1,0.65,1,1), & 10\text{s} \leq t < 20\text{s},  \\ (1,1,1,0), & 20\text{s} \leq t \leq 35\text{s}.  \end{array}\right. 
    \end{aligned}
\end{equation}
We set the desired trajectory as 
\begin{equation}\label{eq:trajectory}
    \begin{aligned}
        \left\{ \begin{array}{l} x_k = \frac{0.3 \cos(2\pi k/N_s)}{1+\sin^2(2\pi k/N_s)} \\ y_k = \frac{0.4\sin(2\pi k/N_s)\cos(2\pi k/N_s)}{1+\sin^2(2\pi k/N_s)} \\ \theta_k = 0 \end{array} \right.  
    \end{aligned} 
\end{equation}
where $k$ is the simulation instant and $N_s$ is the simulation length. We set the sampling interval as $T_s = 0.1$s. The simulation duration is $35$s. The initial posture of the mobile robot is $[x_0, y_0, \theta_0]^T = [10,0,0]^T$. The covariances of process and observation noises in kinematic and dynamic models are set as $\bm{Q}_{\text{kine},k} = 0.0025\bm{I}_{3\times 3}, \bm{R}_{\text{kine},k}=0.01\bm{I}_{3\times 3}, \bm{Q}_{\text{dyna},k}=0.0001\bm{I}_{3\times 3}$ and $\bm{R}_{\text{dyna},k}=0.0004\bm{I}_{3\times 3}$. The fault parameter vector $\Lambda$ is set following the description in Section 2.2. The posterior probability is initialized as $\pi(\bm{G}_{\text{dyna}}^i|\bm{\Xi}_0)= 1/17, i = 1,2, \cdots, 17$. 

\begin{figure}[htbp]
\begin{center}
\includegraphics[width=8.4cm]{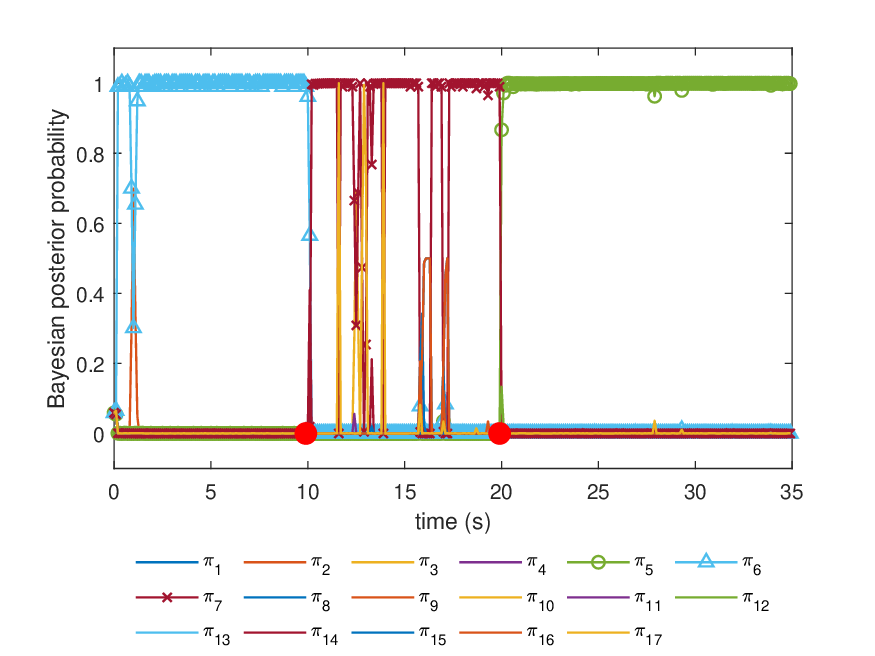}    
\caption{The convergence of posterior probability $\pi(\bm{G}_{\text{dyna}}^i|\bm{\Xi}_k)$ with one actuator in failure, which is represented by $\pi_i$ in the legend. The wheel actuator changes at $10$s and $20$s, marked with red solid circle.} 
\label{fig:onefaultpi}
\end{center}
\end{figure}
Fig.~\ref{fig:onefaultpi} illustrates the real-time learning process of the posterior probabilities $\pi(\bm{G}_{\text{dyna}}^i|\bm{\Xi}_k)$, denoted as $\pi_i$ in the legend. The probability $\pi_6$ converges rapidly to $1$ while all the other probabilities converge to $0$ during the time period $[0\text{s},10\text{s}]$. This means that our learning algorithm identifies that the fault parameter vector $\Lambda^6 = [0.5,1,1,1]$ is the true parameter. The probability $\pi_7$ converge to $1$ after $10\text{s}$, since the fault parameter vector $\Lambda^7 = [1,0.5,1,1]$ is the closest value within the predefined fault set to the actual fault parameter $[1,0.65,1,1]$ shown in (\ref{eq:onefault}). After $20\text{s}$, the probability $\pi$ converged to $1$, correctly learning the true fault parameter as $\Lambda^5 = [1,1,1,0]$, indicating that the actuator $4$ is completely failed. 

\begin{figure}[htbp]
\begin{center}
\includegraphics[width=8.4cm]{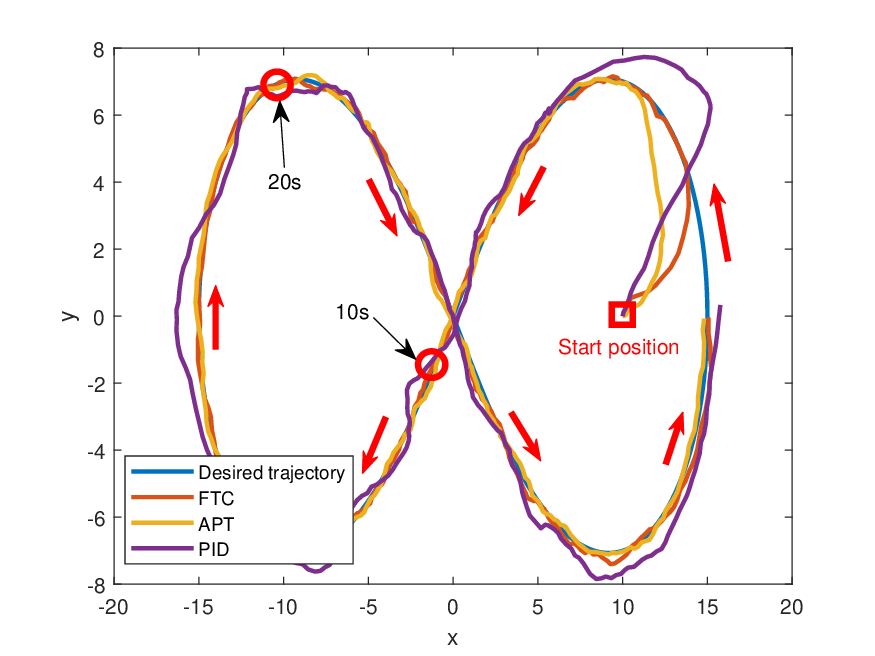}    
\caption{Tracking trajectory of FMWMR with one fault under our developed FTC, adaptive model predictive control (APT) and PID control (PID).} 
\label{fig:compareonefault}
\end{center}
\end{figure}

Fig.~\ref{fig:compareonefault} compares the trajectory tracking performance of our FTC against the PID controller and the adaptive MPC (APT) scheme. Note that we have gone through parameter adjustment for PID and APT before the comparative analysis, and all methods are under the same partial and complete actuator fault sequence shown in (\ref{eq:onefault}).
We observe that both our FTC and the APT methods maintain effective trajectory tracking despite the occurrence of partial or complete actuator faults, demonstrating superior robustness compared to the PID controller. Our FTC tracks the trajectory faster than APT during the starting period, which indicates that our posterior probability learning algorithm exhibits faster convergence compared to the parameter identification by recursive least squares with forgetting factor employed in APT. 

\subsection{Two-fault case}
\begin{figure}[htbp]
\begin{center}
\includegraphics[width=8.4cm]{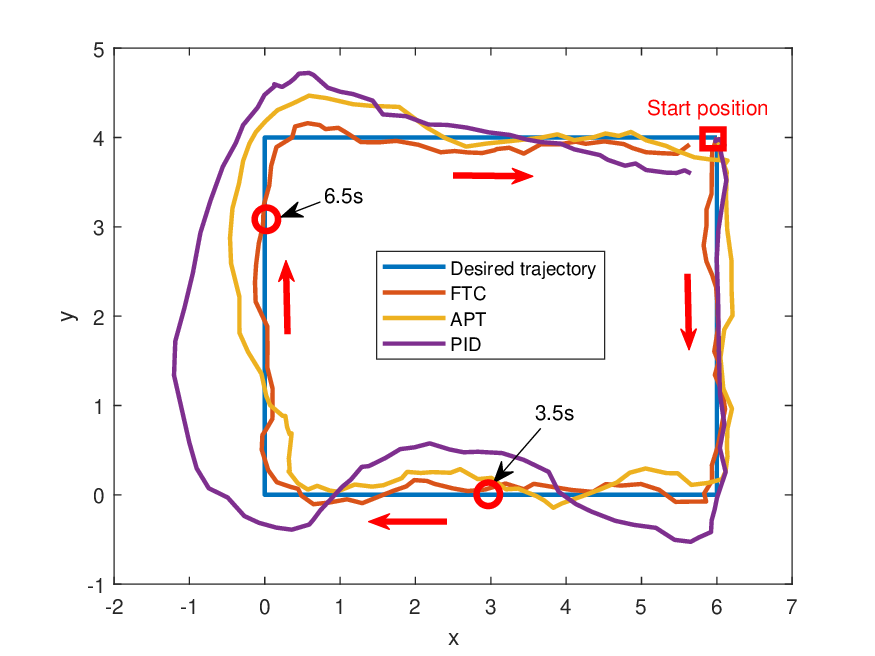}    
\caption{Tracking trajectory of the FMWMR with two actuators in failure under our developed FTC, adaptive model predictive control (APT) and PID control.} 
\label{fig:comparetwofault}
\end{center}
\end{figure}

In this case, we test the trajectory tracking performance for the mobile robot with two-fault occurrences in actuators. The sequential change in actuator faults is set as
\begin{equation}\label{eq:twofault}
	\begin{aligned}
        \Lambda^T (t) = \left\{ \begin{array}{lr} (0.50,0.45,1,1), & 0\text{s} \leq t < 3.5\text{s} ,\\ (1,0,0,1), & 3.5\text{s} \leq t < 6.5\text{s},  \\ (1,1,0.01,0.15), & 6.5\text{s} \leq t \leq 10\text{s}. \end{array}\right. 
    \end{aligned}
\end{equation}
The desired trajectory is set as a square, which is shown in Fig.~\ref{fig:comparetwofault}. The remaining simulation parameters are the same as in the one-fault case. 

\begin{figure}[htbp]
\begin{center}
\includegraphics[width=8.4cm]{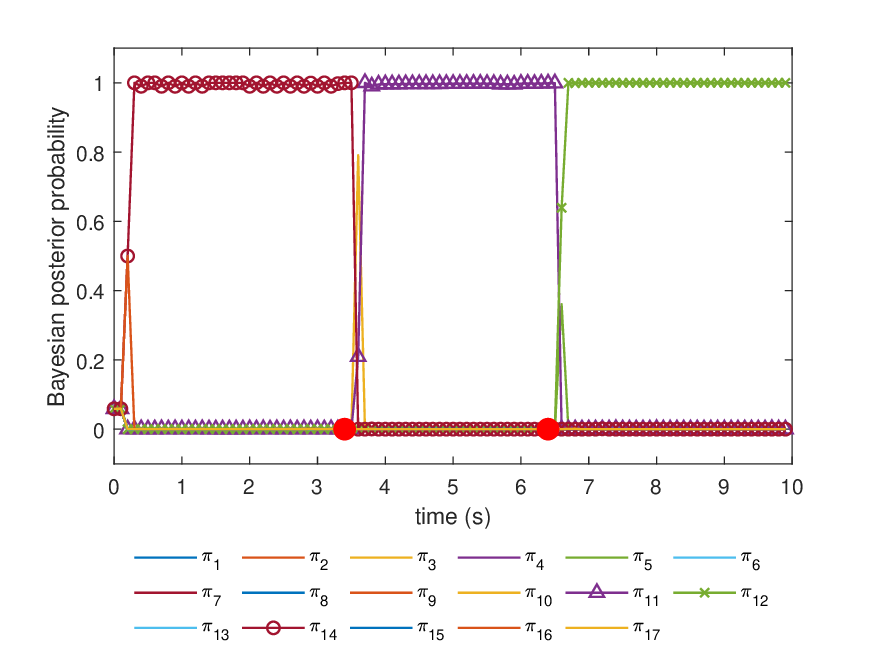}    
\caption{The convergence of posterior probability $\pi(\bm{G}_{\text{dyna}}^i|\bm{\Xi}_k)$ with two actuator faults, which is represented by $\pi_i$ in the legend. The wheel actuator changes at $3.5$s and $6.5$s, marked with red solid circle.} 
\label{fig:twofaultpi}
\end{center}
\end{figure}

Fig.~\ref{fig:twofaultpi} shows the real-time convergence process of posterior probabilities $\pi(\bm{G}_{\text{dyna}}^i|\bm{\Xi}_k)$, labeled as $\pi_i$. The true fault parameter vector is set to $[0.50,0.45,1,1]$ during the initial period, with actuator $1$ in $50\%$ failure and actuator $2$ losing $55\%$ capacity. The probability $\pi_{14}$ converges to $1$ from the starting position, indicating the learned parameter is $\Lambda^{14} = [0.5,0.5,1,1]$, which is the closest value to the true value. The true fault changed to $[1,0,0,1]$ at $3.5$s, with actuator $2$ and $3$ completely failed. We observe that the probability $\pi_{11}$ converges to $1$ while all others converge to $0$ following this fault change, indicating it correctly learns the true fault parameter. After the true fault parameter changed to $[1,1,0.01,0.15]$, the probability $\pi_{12}$ rapidly converges to $1$. This manifests that the learned fault parameter vector is $\Lambda^{12}=[1,1,0,0]$.

Fig.~\ref{fig:comparetwofault} compares the trajectory tracking performance of our FTC against the PID controller and the adaptive MPC (APT) scheme. We observe that our FTC tracks the square better than PID controller and APT method, demonstrating superior robustness and accuracy of tracking performance. 

\subsection{Collision avoidance case}
This simulation tests the collision avoidance capacity of the mobile robot when faults occur. The sequential change in actuator faults is set as
\begin{equation}\label{eq:collision}
	\begin{aligned}
        \Lambda^T (t) = \left\{ \begin{array}{lr} (0.45,1,1,1), & 0\text{s} \leq t < 3\text{s} ,\\ (1,0.65,1,1), & 3.5\text{s} \leq t < 6.5\text{s},  \\ (1,1,0,0), & 6.5\text{s} \leq t \leq 19\text{s}.  \end{array}\right. 
    \end{aligned}
\end{equation}

\begin{figure}[htbp]
\begin{center}
\includegraphics[width=9cm]{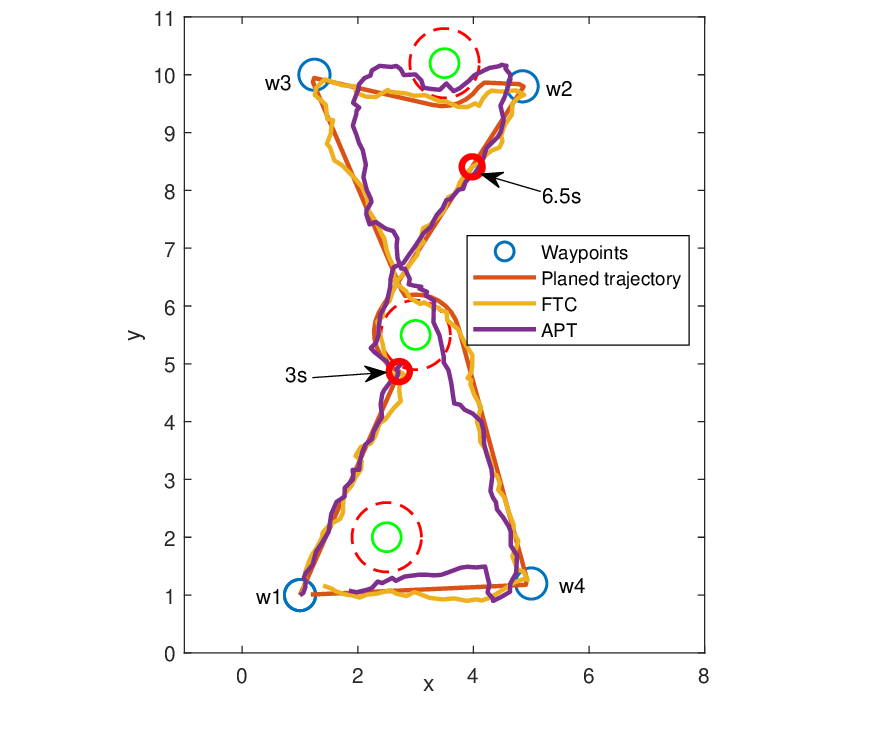}    
\caption{Trace of the FMWMR with three static obstacles in the workspace under our developed FTC and adaptive model predictive control (APT), where he green circles are obstacles, the red dashed circles are danger zones.} 
\label{fig:collisionavoid}
\end{center}
\end{figure}

Fig.~\ref{fig:collisionavoid} compares the trace of FMWMR under our FTC with the APT. The mobile robot under our FTC successfully reaches the desired waypoint (with the sequence w1 $\rightarrow$ w2 $\rightarrow$ w3 $\rightarrow$ w4 $\rightarrow$ w1), and efficiently avoids collision with three obstacles. The APT shows a worse waypoint trace and the mobile robot moves into the dangerous zone that is closer to the obstacle, after $6.5$s when two actuators completely fail. This indicates advanced robustness and safety of our FTC scheme over APT.  

\begin{figure}[htbp]
\begin{center}
\includegraphics[width=8.4cm]{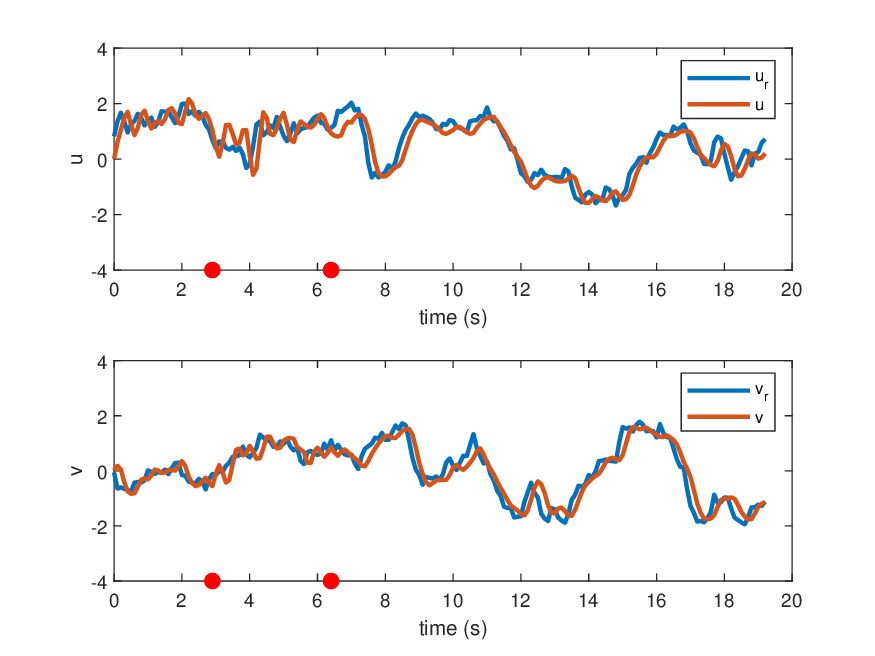}    
\caption{The velocity tracking of the mobile robot under our FTC method. $u_r$ and $v_r$ represent the desired velocity obtained in the kinematics loop.  } 
\label{fig:FTC_velocity}
\end{center}
\end{figure}

\begin{figure}[htbp]
\begin{center}
\includegraphics[width=8.4cm]{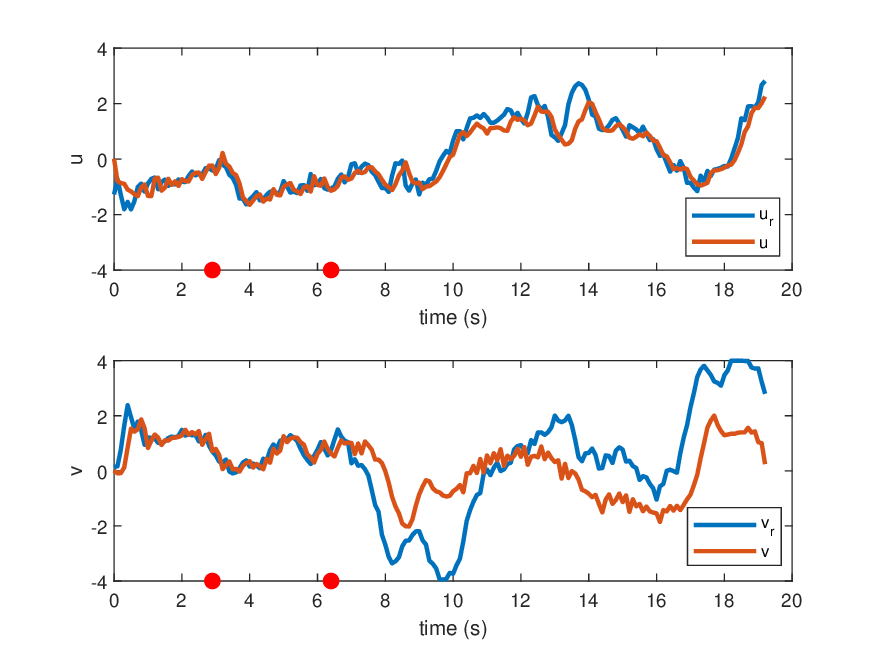}    
\caption{The velocity tracking of the mobile robot under the APT method. $u_r$ and $v_r$ represent the desired velocity obtained in the kinematics loop.  } 
\label{fig:APT_velocity}
\end{center}
\end{figure}

Fig.~\ref{fig:FTC_velocity} and Fig.~\ref{fig:APT_velocity} compare the longitudinal and lateral velocity tracking performance under our FTC and APT. The results show that both our FTC and APT track the reference velocity well with one actuator in partial failure (the period before the second red dot). After two actuators completely fail, our FTC shows better velocity tracking than the APT. This indicates that our FTC algorithm has advanced adaptation and robustness capacity.

\section{Conclusion}\label{sec:5}

This work presents a novel fault tolerant control (FTC) method for FMWMRs. This method ensures robustness and safety for autonomous mobile robots despite varying levels of fault corruption. The computation of our control law is of low complexity, which is friendly to source-limited embedded systems and ideally suited for practical implementation. Furthermore, the presented FTC method is generalizable and can be applied to aerial or underwater vehicles with actuator faults. Our future work will focus on the deployment and experimental validation of our FTC algorithm in real-world Mecanum wheeled mobile robots.

\bibliography{ifacconf}             
                                                   







\end{document}